\pdfoutput=1
\documentclass[12pt]{article}
\usepackage[margin=1in]{geometry}
\usepackage{amsmath,amssymb,amsthm}
\usepackage{graphicx}
\usepackage[hidelinks]{hyperref}
\usepackage{natbib}
\usepackage{microtype}
\usepackage{longtable}
\usepackage{array}
\usepackage{setspace}
\usepackage{ragged2e}
\setcitestyle{aysep={},notesep={:}}
\newtheorem*{relationship}{Key relationship}
\newcommand{\E}{\mathbb{E}}

\title{\vspace*{-1.5cm}\Large\setstretch{1}Can AI Do Formal Demography?\\
Why Vaccinating the Oldest Saves Both the Most Lives\\
and the Most Years of Life}
\author{\Large Joshua R. Goldstein and Kenneth W. Wachter%
\thanks{\linespread{1}\footnotesize Department of Demography, University of California,
Berkeley. Division of labor: Goldstein asked Claude Opus
5.5 (Anthropic) to explain the original proof and then for a simpler
one; pressed for an exhaustive search for precedents; insisted on
notation built on $1/\mu$ and $\mu'/\mu$; simplified the exposition of
Step 2; noticed that the Gompertz bound behaves like an approximation;
asked for an actuarial source for $e' = \mu e - 1$; and chose the
title, framing, and structure. Wachter found the 2021 proof on which
the paper builds; asked whether the lower bound is equivalent to the
convexity of $1/\mu$, which led to the exact form of the condition;
pressed for conditions that can be checked from the hazard alone;
checked the derivation; and reviewed the drafts.
Claude proposed the tangent-line proof and the lower bound, found
Cohen's identity and the reliability-theory precedents, made the
figures, and wrote the draft. The authors are responsible for the
content. See also the ``Use of AI'' section.}}
\date{}

\begin{document}
\RaggedRight
\setlength{\parindent}{1.5em}
\maketitle
\thispagestyle{plain}

\begin{abstract}
\setstretch{1}\noindent
Can AI contribute to formal demography? We report an experiment in which
an AI model, Claude, was asked to explain and then simplify the proof of
an earlier result: that vaccinating the oldest against COVID-19 saves
both the most lives and the most years of life, because the years saved
per person treated rise with age. The model proposed a new proof that is
shorter, more intuitive, and somewhat more general than the original,
built from a known life-table identity and the definition of convexity. It states the condition for the result as the
convexity of the reciprocal of the hazard, which can be checked from the
hazard alone and holds for most familiar parametric models, and it
yields a lower bound on life expectancy whenever that condition is met.
Prompted by a further question, the model also gave an exact form of the
condition, showing how far convexity can be relaxed. The model traced
these identities and related results in earlier literature. An edited
record of the sessions is provided, and complete transcripts are
available on request. The exercise
suggests that AI can revisit established results, make them easier
to understand, and prove valuable generalizations.

\vspace{2.2em}
\noindent\textsc{Keywords}\quad Formal demography, life expectancy, mean
residual life, \mbox{vaccine prioritization}, \mbox{human--AI} collaboration
\end{abstract}

\clearpage

\section*{Introduction}

{\itshape
\noindent AI systems have begun to make discoveries in mathematics. In
September 2026, OpenAI announced a proof, found by a large system of AI
agents and checked in the proof assistant Lean, that solutions of the
three-dimensional Navier--Stokes equations can break down in finite
time, a question posed as one of the Clay Millennium Prize problems
\citep{openai_2026, quanta_2026}. The claim is still under scrutiny,
but it is no longer far-fetched that AI will settle problems that have
resisted mathematicians for generations.

Our aim here is more modest. Formal demography has few famous open
problems of this kind. For a field like ours, AI may instead be most
useful in helping us understand, and even deepen, our own previously
established results. We tested this on a result from our earlier work
with Tom Cassidy: that vaccinating the oldest against COVID-19 saves
both the most lives and the most years of life
\citep{goldstein_cassidy_wachter_2021}. Cassidy and Goldstein had
worked out specific cases, and Wachter found the proof that established
the result in general, by an indirect route. One of us (Goldstein) asked
Claude, an AI model made by Anthropic, to explain that proof and then to
find a simpler one.

What follows is the result, written largely by Claude under our
direction. The new proof is shorter than the original. It restates the
original condition in a more interpretable form, the convexity of the
life expectancy a person would have if the hazard stayed frozen at its
current level. This is a condition on the hazard itself, which matters
because life expectancy rarely has a simple analytic form, and it holds
for most familiar parametric models with increasing hazards. The proof
also yields a lower bound on life expectancy that holds whenever the
condition is met. Then, when the other of us (Wachter), reading the
draft, asked whether that bound was equivalent to the condition, Claude
gave an exact expression for the curvature of life expectancy in terms
of the curvature of the frozen-hazard life expectancy. It shows how far
convexity can be relaxed, although, because it involves the person-years
lived at later ages, it cannot be checked from the hazard alone. The
section on the use of AI at the end describes who did what.

This experiment suggests at least one use for AI in formal demography:
revisiting earlier results, finding arguments that show more clearly
why they hold, and connecting them to the larger body of what is
already known. Here, Claude revisited our 2021 result, found a new, simpler proof
that shows why it holds, and connected it both to an identity
published by \citet{cohen_2010} and to work by reliability theorists.

This example, we believe, offers some general lessons, at least as of
this writing in late 2026. First, AI is already
useful for revisiting what we think we know. Initial proofs are rarely
the clearest, and improving them no longer has to wait for someone with
the time and inclination; it can be done in a few hours of interaction
with an AI model. Second, the gains came from combining human and machine: the
model derived, searched, and drafted, while humans originated the
questions, asked the follow-up that led to the exact condition, and
judged what came back. That division of labor may
not last; models may soon ask more of the follow-up questions
themselves. Even then, humans will need to steer, deciding which of the
many questions a model can generate are the ones worth pursuing. Third, authors revisiting their own results may be particularly
well situated to work this way, because they know where their original
argument was least satisfying and already know enough to judge the value
of what the AI produces.
}

\section*{The Key Relationship}

Let $\mu(x) > 0$ be the force of mortality, $\ell(x)$ survivorship,
$d(x) = \mu(x)\,\ell(x)$ the life table density of deaths, and $X$ the
age at death. Remaining life expectancy at age $x$ (assumed finite) is the mean number of
years still to be lived by those alive at $x$,
\[
e(x) \;=\; \E[\,X - x \mid X > x\,]
\;=\; \frac{1}{\ell(x)} \int_x^\infty \ell(y)\,dy .
\]

\begin{relationship}
If $1/\mu(y)$ is convex for $y \ge x$, then
\begin{equation}
\label{eq:key}
\frac{d}{dx}\,\bigl[\mu(x)\,e(x)\bigr] \;\ge\; 0 .
\end{equation}
\end{relationship}

In words, the product of the hazard and life expectancy rises with age
whenever the reciprocal of the hazard curves upward. Because of the standard
identity $e'(x) = \mu(x)\,e(x) - 1$, this is the same as
saying that life expectancy is convex in age, $e''(x) \ge 0$. If $1/\mu$
is strictly convex, both inequalities are strict. The product stays
constant only in the boundary case where $1/\mu$ is a straight line, flat or sloped,
as with constant mortality or de Moivre's linear survival curve.

The quantity $1/\mu(x)$ has a natural reading. It is the life
expectancy a person aged $x$ would have if the hazard stayed frozen at
its current level. We will call it the \emph{frozen-hazard life
expectancy}. \citet{cohen_2010}, who showed that life expectancy is
``the death-weighted average of the reciprocal of the survival-specific
force of mortality,'' called it the reciprocal of the force of
mortality; we use the shorter name for its interpretation. Under the Gompertz model, $\mu(x) = \alpha e^{\beta x}$,
the frozen-hazard life expectancy $e^{-\beta x}/\alpha$ declines
exponentially. That is a convex function, so the relationship applies at
every age.

Why should anyone care whether $\mu(x)e(x)$ rises? It is the number of
person-years saved by averting a death at age $x$, scaled by the risk
of dying at that age. Consider COVID-19 vaccination. If a vaccine cuts
each recipient's risk of dying by the same proportion at every age, the
expected years of life saved per dose given at age $x$ are proportional
to $\mu(x)e(x)$ \citep{goldstein_cassidy_wachter_2021}. The oldest have
the highest risk of dying but the fewest years left to lose. When
$\mu(x)e(x)$ rises with age, the first effect wins: vaccinating the
oldest first saves not only the most lives but also the most years of
life, and the two goals point in the same direction.

\section*{Proof}

The proof has three steps. Throughout, $\mu'(x)/\mu(x)$ is the rate of
ageing, the relative rate at which the hazard rises with age. Under the
Gompertz model it is the constant $\beta$.

\paragraph{Step 1: Life expectancy is an average of frozen-hazard life expectancies.}
Since $d(y) = \mu(y)\,\ell(y)$, we can write
$\ell(y) = \frac{1}{\mu(y)}\, d(y)$, and so
\begin{equation}
\label{eq:identity}
e(x) \;=\; \frac{1}{\ell(x)} \int_x^\infty \frac{1}{\mu(y)}\, d(y)\,dy
\;=\; \E\!\left[\, \frac{1}{\mu(X)} \;\middle|\; X > x \right].
\end{equation}
In words, actual life expectancy equals
the average frozen-hazard life expectancy, evaluated at the age at which
each survivor eventually dies. The identity needs no assumptions beyond
$\mu > 0$. It appears implicitly in \citet{bradley_gupta_2003} and
explicitly in \citet{cohen_2010}, who wrote it with the surviving
proportion, rather than age, as the variable of integration.

\paragraph{Step 2: Apply the tangent-line inequality.}
A convex function lies on or above each of its tangent lines. The slope
of $1/\mu$ at age $x$ is $-\mu'(x)/\mu(x)^2$, so for every $y \ge x$,
\[
\frac{1}{\mu(y)} \;\ge\; \frac{1}{\mu(x)}
\left[\, 1 - \frac{\mu'(x)}{\mu(x)}\,(y - x) \right].
\]
Now average both sides over the ages at which those alive at $x$ will
die, by multiplying by $d(y)/\ell(x)$ and integrating. Because the
weights are nonnegative, the inequality is preserved:
\[
\underbrace{\int_x^\infty \frac{1}{\mu(y)}\,\frac{d(y)}{\ell(x)}\,dy}_{e(x)}
\;\ge\;
\frac{1}{\mu(x)}\left[\;
\underbrace{\int_x^\infty \frac{d(y)}{\ell(x)}\,dy}_{1}
\;-\; \frac{\mu'(x)}{\mu(x)}
\underbrace{\int_x^\infty (y - x)\,\frac{d(y)}{\ell(x)}\,dy}_{e(x)}
\;\right].
\]
The left integral is $e(x)$ by the identity (\ref{eq:identity}). On the
right, the first integral is one, because everyone alive at $x$
eventually dies, and the second is the mean remaining lifetime, which is
$e(x)$ by definition. Hence
\[
e(x) \;\ge\; \frac{1}{\mu(x)}\left[\, 1 - \frac{\mu'(x)}{\mu(x)}\, e(x) \right],
\]
or, multiplying through by $\mu(x)$,
\begin{equation}
\label{eq:bound}
\mu(x)\,e(x) \;+\; \frac{\mu'(x)}{\mu(x)}\,e(x) \;-\; 1 \;\ge\; 0 .
\end{equation}

\paragraph{Step 3: Recognize the derivative.}
Differentiate $\mu(x)e(x)$ and use the familiar identity
$e'(x) = \mu(x)e(x) - 1$:
\[
\frac{d}{dx}\bigl[\mu(x)\,e(x)\bigr]
\;=\; \mu'(x)\,e(x) + \mu(x)\bigl[\mu(x)e(x) - 1\bigr]
\;=\; \mu(x)\left[\,\mu(x)\,e(x) + \frac{\mu'(x)}{\mu(x)}\,e(x) - 1\right],
\]
which is nonnegative by (\ref{eq:bound}). Because
$\mu(x)e(x) = 1 + e'(x)$, its derivative is $e''(x)$, and the proof is
complete.\qed

\medskip
\noindent The argument never uses the second derivative of $\mu$, nor
does it require $\mu$ to be increasing. Convexity alone supplies the
tangent line. Where $1/\mu$ has a kink, any supporting line will do.

\section*{Related Relationships}

\paragraph{An equivalent condition on the rate of ageing.}
When \citet{goldstein_cassidy_wachter_2021} first proved that
$\mu(x)e(x)$ rises with age, they did not state the condition in terms
of $1/\mu$. Their proof worked with the logarithm of the hazard, and
the condition it required was a restriction on the rate of ageing,
$\mu'/\mu = d\log\mu/dx$, and its derivative. It is not obvious what
that restriction means, or that it is the same as convexity of $1/\mu$.
Differentiating $1/\mu$ twice shows that it is:
\[
\frac{d^2}{dx^2}\,\frac{1}{\mu(x)}
\;=\; \frac{1}{\mu(x)}\left[\left(\frac{\mu'(x)}{\mu(x)}\right)^{2}
- \frac{d}{dx}\,\frac{\mu'(x)}{\mu(x)}\right].
\]
So $1/\mu$ is convex exactly when
\begin{equation}
\label{eq:logcond}
\left(\frac{\mu'(x)}{\mu(x)}\right)^{2} \;>\; \frac{d}{dx}\,\frac{\mu'(x)}{\mu(x)} .
\end{equation}
This is the condition of \citet{goldstein_cassidy_wachter_2021}. In
words, the rate of ageing may rise, but not faster than its own square.
Any model with a positive
rate of ageing that is constant or falling with age satisfies it. That
includes the Gompertz model, Weibull hazards $x^{\beta-1}$ with
$\beta > 1$, and the logistic and gamma--Gompertz models with their
mortality plateaus.

\paragraph{A lower bound on life expectancy.}
Wherever $1/\mu$ is convex at all ages above $x$, inequality
(\ref{eq:bound}) is a bound worth having in its own right:
\[
e(x) \;\ge\; \frac{1}{\mu(x) + \mu'(x)/\mu(x)} .
\]
Life expectancy is at least the reciprocal of the hazard plus the rate
of ageing. Under the Gompertz model the rate of ageing is $\beta$, and
the bound becomes
\begin{equation}
\label{eq:gompbound}
e(x) \;\ge\; \frac{1}{\mu(x) + \beta}.
\end{equation}
The bound becomes exact as the hazard grows large relative to $\beta$,
but it understates life expectancy at the ages where most deaths
occur.\footnote{Under the Gompertz model, $e(x) = e^{z}E_1(z)/\beta$,
where $z = \mu(x)/\beta$ and $E_1$ is the exponential integral. The
bound is then the classical inequality $1/(z+1) < e^{z}E_1(z) < 1/z$
(see the History section), so
$1/(\mu+\beta) < e(x) < 1/\mu$. With $\beta = 0.1$, the lower bound
falls short of the exact value by 22\% at $\mu = 0.07$ (about age
80 in low-mortality countries), 8\% at $\mu = 0.2$, and 2\%
at $\mu = 0.5$.}

\paragraph{The boundary case.}
When $1/\mu$ is linear, the tangent line is exact, so the bound holds
with equality and $\mu(x)e(x)$ is constant. This happens when
$\mu(x) = 1/(a + bx)$ with $b < 1$, the generalized Pareto family. Its
rate of ageing is $-b\,\mu(x)$, and its life expectancy
$e(x) = (a + bx)/(1 - b)$ is linear in age. The family includes two
familiar life tables. De Moivre's linear survival curve,
$\ell(x) = 1 - x/\omega$, is the case $a = \omega$, $b = -1$, with
$e(x) = (\omega - x)/2$ and $\mu(x)e(x) = 1/2$ at every age. Constant
mortality, where $e(x) = 1/\mu$, is the case $b = 0$.

\begin{figure}[tbp]
\centering
\includegraphics[width=\textwidth]{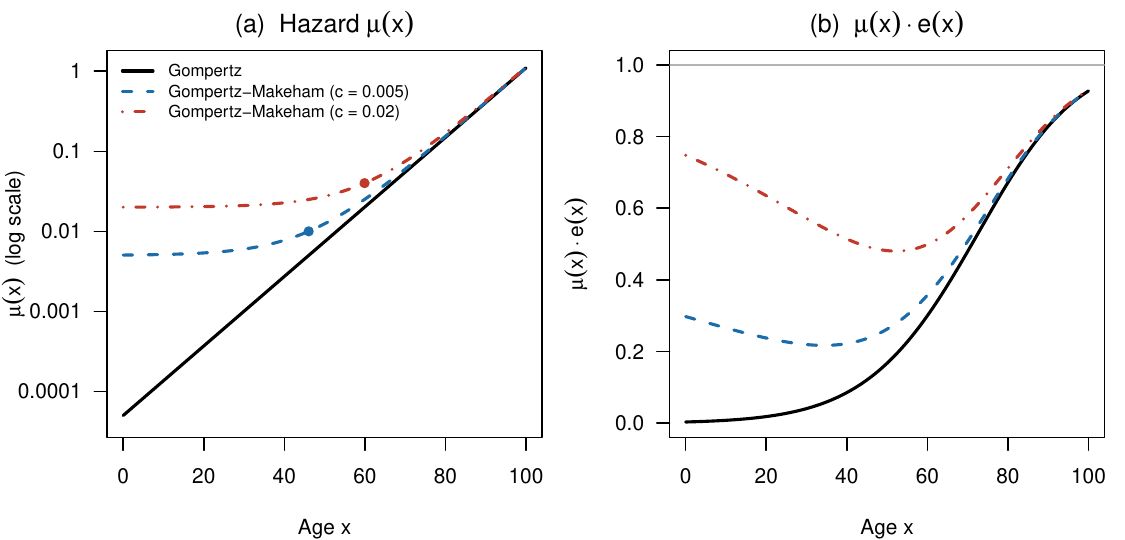}
\caption{(a) Gompertz and Gompertz--Makeham hazards
($\alpha = 5\times 10^{-5}$, $\beta = 0.1$). Dots mark the age
$\log(c/\alpha)/\beta$ above which $1/\mu$ is convex. (b) The product
$\mu(x)e(x)$. It rises at all ages under Gompertz mortality, and at all
ages above the dot under Gompertz--Makeham mortality. All three curves
approach 1 at the oldest ages.}
\label{fig:makeham}
\end{figure}

\paragraph{Where the relationship fails.}
Under Gompertz--Makeham mortality, $\mu(x) = c + \alpha e^{\beta x}$,
\[
\frac{d^2}{dx^2}\,\frac{1}{\mu(x)} = \frac{\beta^{2} A\,(A - c)}{(c + A)^{3}},
\qquad A = \alpha e^{\beta x}.
\]
So $1/\mu$ is convex only above the age where the senescent component
overtakes the background component, $x^{*} = \log(c/\alpha)/\beta$.
Figure~\ref{fig:makeham} shows that this failure is not just a failure
of a sufficient condition. With background mortality of
$c = 0.005$, $\mu(x)e(x)$ actually falls until about age 34. With
$c = 0.02$ it falls until about age 52. More generally, any local
minimum of the hazard, such as the childhood minimum or the trough
after the young-adult accident hump, is a local maximum of $1/\mu$,
where convexity fails.

\paragraph{How far convexity can be relaxed.}
Because $\mu(x)e(x) = 1 + e'(x)$, the product rises exactly where life
expectancy is convex, so an exact condition is always available in
terms of $e$ itself. Keeping, rather than discarding, the gap in Step~2
expresses that condition through the hazard. Without assuming
convexity, the gap between $1/\mu$ and its tangent line at $x$ is the
curvature accumulated between $x$ and $y$:
\[
\frac{1}{\mu(y)} \;-\; \frac{1}{\mu(x)}
\left[\, 1 - \frac{\mu'(x)}{\mu(x)}\,(y - x) \right]
\;=\; \int_x^y (y - t) \left(\frac{1}{\mu}\right)''\!(t)\,dt .
\]
Average both sides over the ages at which those alive at $x$ will die,
as in Step~2, and change the order of integration on the right. The
weight on the curvature at age $t$ becomes
$\int_t^\infty (y - t)\, d(y)\,dy / \ell(x) = T(t)/\ell(x)$, where
$T(t) = \int_t^\infty \ell(y)\,dy$ is the person-years lived above
$t$. With Step~3, this gives
\begin{equation}
\label{eq:exact}
e''(x) \;=\; \frac{d}{dx}\bigl[\mu(x)\,e(x)\bigr]
\;=\; \frac{\mu(x)^2}{\ell(x)} \int_x^\infty \left(\frac{1}{\mu}\right)''\!(t)\; T(t)\,dt .
\end{equation}
So $\mu(x)e(x)$ rises at $x$, and the lower bound holds at $x$, if and
only if the curvature of $1/\mu$ above $x$, weighted by person-years
lived, is nonnegative on balance. Equivalently, $1/\mu$ must lie above
its tangent at $x$ on average over the ages at which survivors die, not
at every age. (At a kink, the curvature includes a point mass equal to
the jump in slope.) Convexity can therefore be relaxed. In the
Gompertz--Makeham example with $c = 0.005$, $\mu(x)e(x)$ begins to rise
at about age 34, although $1/\mu$ is convex only above
$x^{*} \approx 46$. But Equation (\ref{eq:exact}) involves $T$, which,
like $e$, rarely has a simple analytic form, so it cannot be checked
from the hazard alone. Convexity of $1/\mu$ remains the practical test.

\section*{History}

The result offers an alternative to the original proof of
\citet{goldstein_cassidy_wachter_2021}, who showed that $\mu(x)e(x)$
increases with age under condition (\ref{eq:logcond}). Their argument
changed variables from age to cumulative hazard, which turns
$\mu(x)e(x)$ into an average of hazard ratios, and then used a
log-concavity argument. The proof given here is shorter and extends
theirs in two ways. It requires only that $1/\mu$ be convex, so the
hazard need not be twice differentiable and may have kinks. And it
yields the lower bound on life expectancy as a by-product.

The proof combines the identity (\ref{eq:identity}) of \citet{cohen_2010}
with a tangent line. Cohen also derived a lower bound from his identity,
by applying Jensen's inequality to the reciprocal function: the average
of $1/\mu$ over deaths is at least the reciprocal of the average of
$\mu$. That bound differs from ours, which uses the shape of $1/\mu$
across ages, and Cohen did not examine the shape of $e(x)$. We have not found the convexity result or the general lower
bound stated elsewhere, and we would welcome earlier references.

Before \citet{goldstein_cassidy_wachter_2021}, reliability theorists
had made partial statements about the product of the hazard and the
mean residual life, as they call $e(x)$. \citet{kupka_loo_1989} named
$\mu(x)e(x)$ the ``vitality rate.'' They noted that its increase is
equivalent to convexity of $e(x)$, as follows from $e' = \mu e - 1$, and showed that an increasing hazard
alone does not imply it, but they gave no condition on the hazard that
does. \citet[ Theorem 5.3]{gupta_kirmani_2000} showed that a concave
hazard implies a convex mean residual life. Because a positive concave
function has a convex reciprocal, their result is a special case of the
one here. It does not cover the Gompertz model, whose hazard is convex.
\citet{bradley_gupta_2003} studied the limit of $\mu(x)e(x)$ at old
ages in terms of $1/\mu$ and its derivatives, but not its shape at
younger ages. In doing so they wrote life expectancy, in effect, as the
average of $1/\mu$ over ages at death, the identity (\ref{eq:identity})
later stated explicitly by \citet{cohen_2010}, and obtained by
integration by parts its first-derivative counterpart,
$e'(x) = \frac{\mu(x)}{\ell(x)} \int_x^\infty (1/\mu)'(t)\,\ell(t)\,dt$
\citep[ proof of Theorem 4]{bradley_gupta_2003}. One
further integration by parts gives Equation (\ref{eq:exact}). The
condition it expresses, convexity of $e(x)$, was known
\citep{kupka_loo_1989}; we have not found Equation (\ref{eq:exact}),
which states that condition through the curvature of $1/\mu$, elsewhere.

The ingredients are standard. The identity $e'(x) = \mu(x)e(x) - 1$ is
long-standing actuarial knowledge \citep[e.g.,][212]{spurgeon_1922}.
In the Gompertz case, the lower bound (\ref{eq:gompbound}) is a
classical inequality for the exponential integral
\citep[ formula 5.1.19]{abramowitz_stegun_1964}.

\section*{Application}

The result was motivated by COVID-19 vaccine allocation.
\citet{goldstein_cassidy_wachter_2021} observed that if vaccine efficacy
$\theta$ is constant over age and the disease kills at a constant
fraction $\rho$ of all-cause mortality, the expected years of life saved
per dose given at age $x$ are $\theta \rho\, \mu(x)\, e(x)$. The key
relationship says that, wherever mortality is Gompertz-like, this
quantity is largest for the oldest. Prioritizing the oldest therefore
saves both the most lives and the most years of life. This was the case
for COVID-19 in the United States, Germany, and South Korea
\citep{goldstein_cassidy_wachter_2021}. Where background mortality is
high, or where the disease's age pattern departs from that of all-cause
mortality, the conclusion need not hold at every age
\citep{pifarre_etal_2021}. Even then, as Figure~\ref{fig:makeham}
shows, the oldest may still save the most years.

The theorem also clarifies a familiar demographic result. A small
proportional reduction in mortality at age $x$ raises life expectancy at
birth by an amount proportional to $\ell(x)\mu(x)e(x)$
\citep[see, e.g.,][]{vaupel_1986}. These weights typically decline at
the oldest ages, which can suggest that mortality reductions there are
of little value. The decline comes entirely from the factor $\ell(x)$,
the number of people who survive to age $x$. The remaining factor,
$\mu(x)e(x)$, is the gain per person alive at $x$, and by the key
relationship it rises with age. Gains summed over a birth cohort and
gains per person treated therefore have opposite age gradients, and it
is the second that matters when deciding whom to treat.

The same logic applies to other interventions that remove a fixed share
of each person's risk of death, at once and at the same cost at every
age, and that leave those saved facing the ordinary life table
afterwards. Many age-prioritization questions are more complicated
because one of these conditions fails. Cancer screening and preventive
drugs pay off only after a lag, and the chance of surviving the lag
falls with age. Surgery and intensive care carry risks of their own that
rise with age, so the cost of treatment is not constant. And a
transplanted organ is a durable resource whose working life can outlast
an older recipient, so the benefit depends on life expectancy with and
without the organ, not on $\mu(x)e(x)$. In these cases a real trade-off
between saving lives and saving years of life can arise.

\section*{Use of AI}

{\itshape
\noindent The original work was done in a single interactive session with Claude
Opus 5.5 (Anthropic), run in Claude Code, which let the model read
files, run code, compile \LaTeX, and search the web and, through Goldstein's
library login, licensed full texts. The session took about two and a
half hours on the evening of October 9, 2026, with a brief follow-up
the next morning, and comprised 60 prompts. Goldstein began by asking Claude to
explain the proof at the end of \citet{goldstein_cassidy_wachter_2021}
and to search the reliability literature for precedents. In reporting
on that search, Claude observed that the result is equivalent to the
convexity of life expectancy, and when he asked \texttt{what's your simple
convexity proof}, it proposed the tangent-line argument of the Proof
section. The remaining prompts asked Claude to write the proof up for a
general demographic audience, to check its originality more
exhaustively, and to revise the notation and the exposition of Step 2;
most were short editorial instructions. The sixteen prompts that shaped
the content are listed verbatim in Appendix Table~\ref{tab:prompts}. An
edited record of this session, with prompts verbatim and replies
excerpted, is provided as a supplementary file; the complete transcripts
of all sessions are available from the authors. Goldstein used Claude again, in sessions
beginning on October 22, 2026, to revise the paper for this journal,
including the title, the introduction, and this section. After reading
the draft, Wachter asked whether the lower bound, holding at all ages,
is equivalent to the convexity of $1/\mu$; in reply, Claude showed that
it is not and derived the exact form (\ref{eq:exact}) given in the
Related Relationships section. At Goldstein's request, Claude then
checked every reference against bibliographic databases and the full
texts. It confirmed that all the cited works exist and that the
citations were correct, including a formula number that Claude had
given from memory, with two exceptions, both from the AI sessions. The
publisher of \citet{spurgeon_1922} was wrong. And an article by Keyfitz,
cited with \citet{vaupel_1986} for the effect of a change in mortality at
one age, was one Claude had not read; the result could be confirmed
only in Vaupel, so the citation was dropped.
The division of labor between Claude and the two of us is described in the
footnote on the title page.
}

\section*{Acknowledgments}

As described in the ``Use of AI'' section, this paper
was produced with the assistance of Claude Opus 5.5 (Anthropic).
This work was supported by the \emph{Eunice Kennedy Shriver} National
Institute of Child Health and Human Development (NICHD) of the National
Institutes of Health under award R25HD083136 (Workshops in Formal
Demography).

\clearpage
\appendix
\section*{Appendix}
\renewcommand{\thetable}{A\arabic{table}}

\begin{longtable}{@{}r >{\ttfamily\footnotesize\raggedright\arraybackslash}p{0.86\textwidth}@{}}
\caption{The sixteen prompts that shaped the content, verbatim, in
order. Numbers give each prompt's position among the 60 in the
session. The other prompts were editorial or practical (``open in
skim,'' ``recompile'').}\label{tab:prompts}\\
\hline
\# & \normalfont Prompt \\
\hline
\endfirsthead
\hline
\# & \normalfont Prompt \\
\hline
\endhead
\hline
\endfoot
\multicolumn{2}{@{}l}{\emph{Understanding the original proof}}\\
1 & read the paper by goldstein cassidy and wachter in pnas it should be on disk. and explain the proof to me that is at the end \\
2 & how would you describe the approach of this proof \\[3pt]
\multicolumn{2}{@{}l}{\emph{Precedents}}\\
3 & search the reliability literature for precedents \\[3pt]
\multicolumn{2}{@{}l}{\emph{The new proof}}\\
4 & ok so what's your simple convexity proof. please don't use latex when showing me math on console \\[3pt]
\multicolumn{2}{@{}l}{\emph{Writing it up}}\\
7 & cool. write up a pdf version of the proof. the proof should be accessible to general demogrphaic audience. you can look at goldstein and vaupel's Formal Relationship articles in Demogrpahic Research for style and voice \\[3pt]
\multicolumn{2}{@{}l}{\emph{The lower bound}}\\
9 & wow. your gompertz life expectancy at old ages inequality looks almost like a good approximation, right? \\[3pt]
\multicolumn{2}{@{}l}{\emph{Checking originality}}\\
13 & how could one do a more exhaustive lit check for originality of this result? \\
21 & also look for citers of goldstein, cassidy and wachter to see if someone has come up with different math \\[3pt]
\multicolumn{2}{@{}l}{\emph{Notation}}\\
23 & is s(x) really best notation here. seems like it should be a junior version of S(x) \ldots \\
30 & why did we reject h\^{}\{-1\} \\
31 & rewrite the draft with 1/$\mu$ and $\mu$'/$\mu$, no new symbol \\[3pt]
\multicolumn{2}{@{}l}{\emph{Exposition of Step 2}}\\
39 & oh, maybe flesh out the math of expectations at the cost of a bit more text. when you take expectaitons in part 2 of proof, it's not obvious without paper and pencil what's going on. \\
43 & for me step 2 would be simpler if you just said something like you did at first, averaging \ldots\ but then you actually write down the integrals that are the averages rather than explain everythign in words. also is it ovious that E(y-x) = E(x)? maybe you don't have to do integration by parts. maybe there's a simple logicla reason for this \\[3pt]
\multicolumn{2}{@{}l}{\emph{A source for $e'(x) = \mu(x)e(x) - 1$}}\\
46 & id don't believe the results from calabria and pucini are foundational. this must have been known much, much earlier. the keyfitz an dvaupel stuff seems ok, but probably also restatements. \\
47 & old actuarial knowledge needs a citation, no? \\[3pt]
\multicolumn{2}{@{}l}{\emph{Why the proof is simple}}\\
58 & one thing that is good about the new proof is that it uses life table identities and very simple calculus. just the definition of convexity. no change of variable, for example. am i right in seeing that it is so simple \\
\end{longtable}

\end{document}